\documentclass[wcp]{jmlr}

\usepackage{longtable}

\usepackage{booktabs}

\makeatletter
\let\Ginclude@graphics\@org@Ginclude@graphics 
\makeatother

\jmlrvolume{157}
\jmlryear{2021}
\jmlrworkshop{ACML 2021}

\usepackage{hyperref}
\usepackage{url}
\usepackage{graphicx}
\usepackage{caption}
\usepackage{amsfonts}
\usepackage{amsmath}
\usepackage{multirow}
\usepackage{pgfplots}
\usepackage{pgfplotstable} 
\pgfplotsset{compat=newest}
\usepackage{enumitem}
\usepackage{balance}
\usepackage{wrapfig}
\usepackage{arydshln}

\title{Quaternion Graph Neural Networks}

  \author{\Name{Dai Quoc Nguyen} \Email{dai.nguyen@monash.edu}\\
  \addr Department of Data Science and AI, Monash University, Australia
  \AND
  \Name{Tu Dinh Nguyen} \Email{v.tund21@vinai.io}\\
  \addr VinAI Research, Vietnam
  \AND
  \Name{Dinh Phung} \Email{dinh.phung@monash.edu}\\
  \addr Department of Data Science and AI, Monash University, Australia
 }

\editors{Vineeth N Balasubramanian and Ivor Tsang}

\begin{document}

\maketitle

\begin{abstract}

Recently, graph neural networks (GNNs) have become an important and active research direction in deep learning. It is worth noting that most of the existing GNN-based methods learn graph representations within the Euclidean vector space. Beyond the Euclidean space, learning representation and embeddings in hyper-complex space have also shown to be a promising and effective approach. To this end, we propose Quaternion Graph Neural Networks (QGNN) to learn graph representations within the Quaternion space. As demonstrated, the Quaternion space, a hyper-complex vector space, provides highly meaningful computations and analogical calculus through Hamilton product compared to the Euclidean and complex vector spaces. Our QGNN obtains state-of-the-art results on a range of benchmark datasets for graph classification and node classification. Besides, regarding knowledge graphs, our QGNN-based embedding model achieves state-of-the-art results on three new and challenging benchmark datasets for knowledge graph completion. Our code is available at: \url{https://github.com/daiquocnguyen/QGNN}.

\end{abstract}



\section{Introduction}

Graph representation learning has recently emerged as a new promising learning paradigm for graph-structured data \citep{hamilton2017representation,wu2019comprehensive}, where the goal is to learn a parametric mapping function that embeds the nodes, the subgraphs, or the entire graph into low-dimensional continuous vector spaces; hence the learned vectors can be useful for downstream tasks.
Recently, graph neural networks (GNNs) become an essential strand to learn graph representations \citep{zhang2020network,Nguyen2021Thesis}.
In general, GNNs utilize an aggregation function to update the vector representation of each node by aggregating those of its neighbors \citep{kipf2017semi,velickovic2018graph,Nguyen2019U2GNN}.
GNNs also use a graph-level readout function such as a simple sum pooling to obtain graph embeddings.
As a result, GNNs achieve state-of-the-art performances for downstream tasks such as node classification and graph classification \citep{hamilton2017inductive,xu2019powerful}.

It is noted that most of the existing GNNs learn graph representations 
within the Euclidean space.
Recently, learning representation and embeddings in hyper-complex space have shown to be a promising and effective approach to move beyond the Euclidean space.
For example, some Quaternion space-based methods have been applied in image classification \citep{gaudet2018deep,zhu2018quaternion}, speech recognition \citep{parcolletquaternioncnn,parcollet2019quaternion}, knowledge graph \citep{zhang2019quaternion,Nguyen2020QuatRE}, and machine translation \citep{tay2019lightweight}. 
As the Quaternion space allows highly expressive computations through Hamilton product compared to the Euclidean and complex vector spaces \citep{parcollet2019survey}, we introduce Quaternion Graph Neural Networks (QGNN) 
to learn graph representations within the Quaternion space.
Our proposed model achieves effective performances through extensive experimental evaluation, benchmarking with current state-of-the-art methods, and post-analysis to demonstrate the merits of the proposed methodology, on a wide range of benchmark datasets for the tasks of node classification, graph classification, and knowledge graph completion.
In summary, our contributions can be highlighted as follows:


$\bullet$ We propose Quaternion Graph Neural Networks (QGNN) to learn quaternion embeddings for graph-structured data. To the best of our knowledge, {our work is the first to investigate quaternion embeddings for general graphs with diverse and different structures}.

$\bullet$ QGNN produces state-of-the-art accuracies on citation network datasets for node classification as well as social network and bioinformatics datasets for graph classification.

$\bullet$ Regarding knowledge graphs, we present a new, simple yet effective method that utilizes vanilla GNNs directly on a single undirected graph of entities constructed from a given knowledge graph, followed by a score function to compute the triple score. 
Our QGNN-based knowledge graph embedding model, called SimQGNN, outperforms the existing models and obtains state-of-the-art results on three new and challenging benchmark datasets CoDEx-S, CoDEx-M, and CoDEx-L \citep{safavi2020codex} for knowledge graph completion.\footnote{We propose Gated Quaternion Graph Neural Networks (GQGNN) in Appendix \ref{app:appendixgqnn}.}



\section{Related work} 
\label{sec:relatedwork}

There have been many designs for the aggregation functions proposed in recent literature. The widely-used one is introduced in Graph Convolutional Networks (GCNs) \citep{kipf2017semi}.
Furthermore, Simple Graph Convolution \citep{wu2019simplifying} is a simplified variant of GCN without using the non-linear activation function $\mathsf{g}$.
Graph Isomorphism Network \citep{xu2019powerful} is proposed to leverage an aggregation function based on a multi-layer perceptron (MLP) network of two fully-connected layers.
Besides, Graph Attention Network \citep{velickovic2018graph} extends GCN to compute edge weights following the standard attention technique \citep{bahdanau2014neural} and then employs the multi-head attention technique \citep{vaswani2017attention} to further stabilise the learning process, except the final (prediction) layer where GAT uses averaging.
UGformer \citep{Nguyen2019U2GNN} adapts the transformer self-attention network \citep{vaswani2017attention,dehghani2018universal} to construct an advanced aggregation function to learn node and graph representations.
Hyperbolic Graph Convolutional Neural Networks (HGCN) \citep{chami2019hyperbolic} leverages the expressiveness of GCNs and hyperbolic geometry to map Euclidean input features to embeddings in hyperbolic spaces to learn node representations.
Moreover, Gated GNNs \citep{li2015gated} adopts GRUs \citep{cho2014learning}, unrolls the recurrence for a fixed number of timesteps, and removes the need to constrain parameters to ensure convergence.


Following \citep{xu2018representation,xu2019powerful}, we also employ a concatenation over the vector representations of node $\mathsf{v}$ at the different layers to construct a final vector representation $\boldsymbol{\mathsf{e}}_\mathsf{v}$ for each node $\mathsf{v} \in \mathcal{G}$.
The graph-level readout function can be a simple sum pooling or a complex pooling such as sort pooling \citep{zhang2018end}, hierarchical pooling \citep{cangea2018towards}, and differentiable pooling \citep{Ying2018diffpool}. 
As the sum pooling produces competitive results \citep{xu2019powerful}, we use the simple sum pooling to obtain the embedding $\boldsymbol{\mathsf{e}}_{\mathcal{G}}$ of the entire graph $\mathcal{G}$.

\section{Quaternion background}
\label{sec:quabackground}


A quaternion $q \in \mathbb{H}$ is a hyper-complex number consisting of one real and three separate imaginary components \citep{hamilton1844ii} defined as:
$q = q_r + q_i\boldsymbol{\mathsf{i}} + q_j\boldsymbol{\mathsf{j}} + q_k\boldsymbol{\mathsf{k}}$,
where $q_r, q_i, q_j, q_k \in \mathbb{R}$, and $\boldsymbol{\mathsf{i}}, \boldsymbol{\mathsf{j}}, \boldsymbol{\mathsf{k}}$ are imaginary units that $\boldsymbol{\mathsf{i}}^2 = \boldsymbol{\mathsf{j}}^2 = \boldsymbol{\mathsf{k}}^2 = \boldsymbol{\mathsf{i}}\boldsymbol{\mathsf{j}}\boldsymbol{\mathsf{k}} = -1$.
The operations for the Quaternion algebra are defined as follows:

\textbf{Addition.} 
The addition of two quaternions $q$ and $p$ is defined as:
$q + p = (q_r + p_r) + (q_i + p_i) \boldsymbol{\mathsf{i}} + (q_j + p_j)\boldsymbol{\mathsf{j}} + (q_k + p_k)\boldsymbol{\mathsf{k}}$

\textbf{Norm.} The norm $\|q\|$ of a quaternion $q$ is computed as:
$\|q\| = \sqrt{q_r^2 + q_i^2 + q_j^2 + q_k^2}$.
And the normalized or unit quaternion $q^\triangleleft$ is defined as:
$q^\triangleleft = \frac{q}{\|q\|}$

\textbf{Scalar multiplication.} 
The multiplication of a scalar $\lambda$ and $q$ is computed as:
$\lambda q = \lambda q_r + \lambda q_i\boldsymbol{\mathsf{i}} + \lambda q_j\boldsymbol{\mathsf{j}} + \lambda q_k\boldsymbol{\mathsf{k}}$


\textbf{Hamilton product.} The Hamilton product $\otimes$ (i.e., the quaternion multiplication) of two quaternions $q$ and $p$ is defined as:
\begin{eqnarray}
q \otimes p &=& (q_r p_r - q_i p_i - q_j p_j - q_k p_k) + (q_i p_r + q_r p_i - q_k p_j + q_j p_k)\boldsymbol{\mathsf{i}} \nonumber \\
&+& (q_j p_r + q_k p_i + q_r p_j - q_i p_k)\boldsymbol{\mathsf{j}} + (q_k p_r - q_j p_i + q_i p_j + q_r p_k)\boldsymbol{\mathsf{k}} 
\label{equa:halproduct}
\end{eqnarray}
We can express the Hamilton product of $q$ and $p$ in the following form:
\begin{equation}
q \otimes p = 
\begin{bmatrix}
1\\
\boldsymbol{\mathsf{i}}\\
\boldsymbol{\mathsf{j}}\\
\boldsymbol{\mathsf{k}}
\end{bmatrix}^\top
\begin{bmatrix}
q_r & -q_i & -q_j & -q_k\\
q_i & q_r & -q_k & q_j\\
q_j & q_k & q_r & -q_i\\
q_k & -q_j & q_i & q_r
\end{bmatrix}
\begin{bmatrix}
p_r\\
p_i\\
p_j\\
p_k
\end{bmatrix}
\label{equa:halproduct_newform}
\end{equation}
We note that the Hamilton product is not commutative, i.e., $q \otimes p \neq p \otimes q$.

\textbf{Quaternion-inner product.}  The quaternion-inner product $\bullet$ of two quaternion vectors $\boldsymbol{q}$ and $\boldsymbol{p} \in \mathbb{H}^n$ returns a scalar as:
\begin{equation}
\boldsymbol{q} \bullet \boldsymbol{p} = \boldsymbol{q}_{r}^\textsf{T}\boldsymbol{p}_{r} + \boldsymbol{q}_{i}^\textsf{T}\boldsymbol{p}_{i} + \boldsymbol{q}_{j}^\textsf{T}\boldsymbol{p}_{j} + \boldsymbol{q}_{k}^\textsf{T}\boldsymbol{p}_{k}
\end{equation}

\textbf{Concatenation.} We define a concatenation of two quaternion vectors $\boldsymbol{q}$ and $\boldsymbol{p}$ as:
\begin{equation}
\left[\boldsymbol{q} \parallel \boldsymbol{p}\right] = \left[\boldsymbol{q}_r \parallel \boldsymbol{p}_r\right] + \left[\boldsymbol{q}_i \parallel \boldsymbol{p}_i\right]\boldsymbol{\mathsf{i}} + \left[\boldsymbol{q}_j \parallel \boldsymbol{p}_j\right]\boldsymbol{\mathsf{j}} + \left[\boldsymbol{q}_k \parallel \boldsymbol{p}_k\right]\boldsymbol{\mathsf{k}} 
\end{equation}

\section{Graph representation learning within the Quaternion space}

Most of the existing GNNs focus on learning graph representations within the Euclidean space.
We, however, note that learning representation and embeddings in hyper-complex space have recently shown to be a promising and effective approach to move beyond the Euclidean space \citep{parcollet2019survey}.
This motivates us to move beyond the Euclidean space and consider the Quaternion space -- a hyper-complex vector space -- which provides highly expressive computations through the Hamilton product. 

\subsection{Quaternion Graph Neural Networks (QGNN)}
\label{sec:ourmodel}


\begin{figure}[!ht]
\centering
\includegraphics[width=0.65\textwidth]{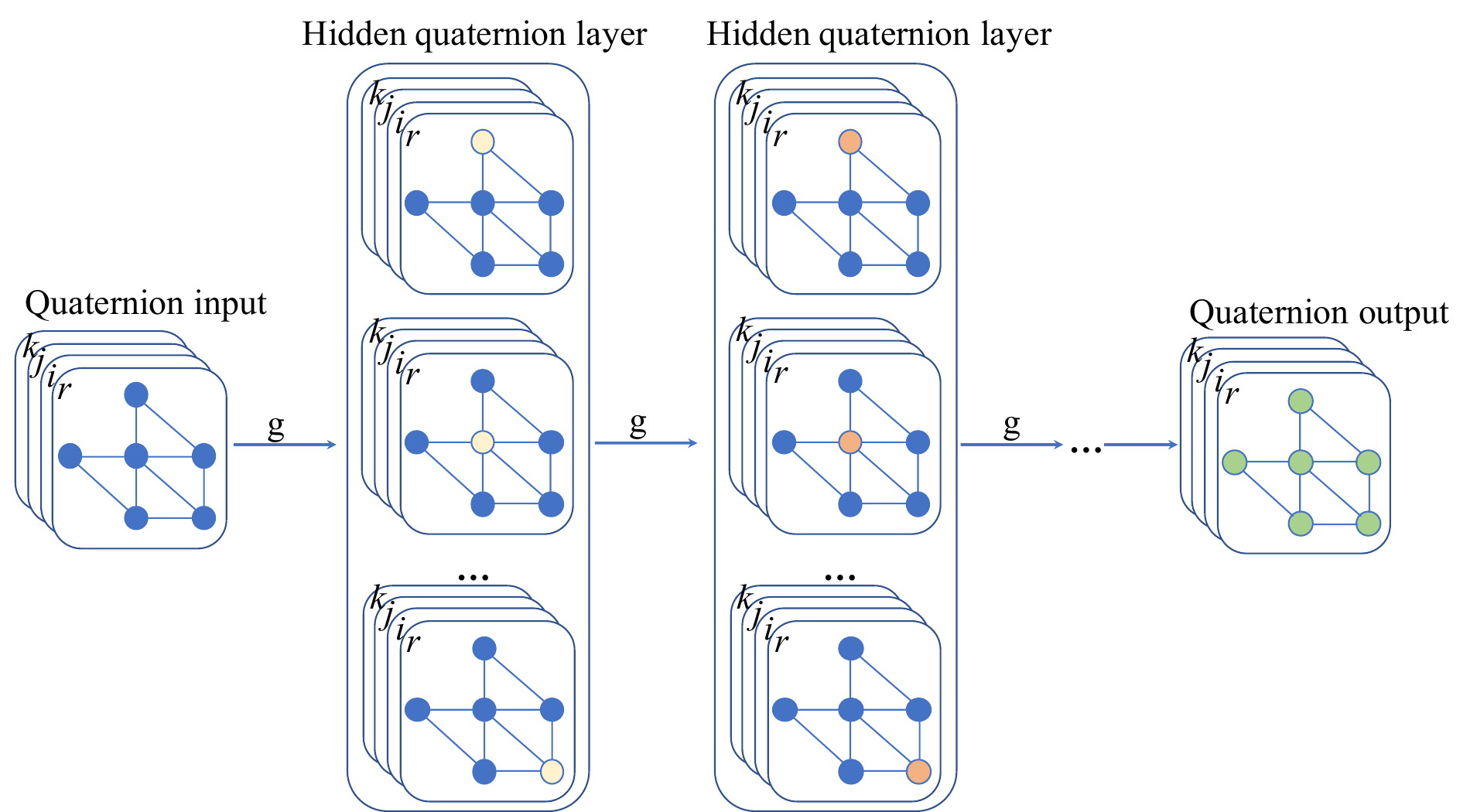}
\captionof{figure}{Illustration of our QGNN.}
\label{fig:QGNN_architecture}
\end{figure}

\noindent We propose Quaternion Graph Neural Networks (QGNN) to learn quaternion embeddings for graph-structured data.
Our QGNN can be seen as a generalization of GCN within the Quaternion space.
We represent each graph $\mathcal{G} = \left(\mathcal{V}, \mathit{E}, \{\boldsymbol{\mathsf{h}}_\mathsf{v}^{(0),Q}\}_{\mathsf{v} \in \mathcal{V}}\right)$, where $\mathcal{V}$ is a set of nodes, $\mathit{E}$ is a set of edges.
In particular, 
the aggregation function in QGNN is defined as:\footnote{In practice, we also implement QGNN efficiently using sparse matrix multiplications.}
\begin{equation}
\boldsymbol{\mathsf{h}}_{\mathsf{v}}^{(l+1),Q} = \mathsf{g}\left(\sum_{\mathsf{u} \in \mathcal{N}_\mathsf{v}\cup\left\{\mathsf{v}\right\}}a_{\mathsf{v},\mathsf{u}}\textbf{W}^{(l),Q}\otimes\boldsymbol{\mathsf{h}}_{\mathsf{u}}^{(l),Q}\right) , \forall \mathsf{v} \in \mathcal{V}
\label{equa:QGNN}
\end{equation}
where $\mathcal{N}_\mathsf{v}$ is the set of neighbors of node $\mathsf{v}$; $a_{\mathsf{v},\mathsf{u}}$ is an edge constant between nodes $\mathsf{v}$ and $\mathsf{u}$ in the re-normalized adjacency matrix $\tilde{\textbf{D}}^{-\frac{1}{2}}\tilde{\textbf{A}}\tilde{\textbf{D}}^{-\frac{1}{2}}$, wherein $\tilde{\textbf{A}} = \textbf{A} + \textbf{I}$ where $\textbf{A}$ is the adjacency matrix, $\textbf{I}$ is the identity matrix, and  $\tilde{\textbf{D}}$ is the diagonal node degree matrix of $\tilde{\textbf{A}}$. We use the superscript $^Q$ to denote the Quaternion space; $\textbf{W}^{(l),Q}$ is a quaternion weight matrix; $\boldsymbol{\mathsf{h}}_\mathsf{v}^{(0),Q}$ is the quaternion feature vector of node $\mathsf{v}$;  and $\mathsf{g}$ is a nonlinear activation function (such as $\mathsf{ReLU}$ or $\mathsf{tanh}$) and can be adopted to each quaternion element \citep{parcollet2019quaternion} as:
\begin{equation}
\mathsf{g}(q) = \mathsf{g}(q_r) + \mathsf{g}(q_i)\boldsymbol{\mathsf{i}} + \mathsf{g}(q_j)\boldsymbol{\mathsf{j}} + \mathsf{g}(q_k)\boldsymbol{\mathsf{k}}
\end{equation}

Correspondingly, we represent the quaternion vector $\boldsymbol{\mathsf{h}}_{\mathsf{u}}^{(l),Q} \in \mathbb{H}^n$ and the quaternion weight matrix $\textbf{W}^{(l),Q} \in \mathbb{H}^{m\times n}$ as:
\begin{equation}
\boldsymbol{\mathsf{h}}_{\mathsf{u}}^{(l),Q} =  \boldsymbol{\mathsf{h}}_{\mathsf{u},r}^{(l)} + \boldsymbol{\mathsf{h}}_{\mathsf{u},i}^{(l)}\boldsymbol{\mathsf{i}} + \boldsymbol{\mathsf{h}}_{\mathsf{u},j}^{(l)}\boldsymbol{\mathsf{j}} + \boldsymbol{\mathsf{h}}_{\mathsf{u},k}^{(l)}\boldsymbol{\mathsf{k}} 
\end{equation}
\begin{equation}
\textbf{W}^{(l),Q} =  \textbf{W}_r^{(l)} + \textbf{W}_i^{(l)}\boldsymbol{\mathsf{i}} + \textbf{W}_j^{(l)}\boldsymbol{\mathsf{j}} + \textbf{W}_k^{(l)}\boldsymbol{\mathsf{k}}
\end{equation}
where $\boldsymbol{\mathsf{h}}_{\mathsf{u},r}^{(l)}$, $\boldsymbol{\mathsf{h}}_{\mathsf{u},i}^{(l)}$, $\boldsymbol{\mathsf{h}}_{\mathsf{u},j}^{(l)}$, and $\boldsymbol{\mathsf{h}}_{\mathsf{u},k}^{(l)} \in \mathbb{R}^n$; and $\textbf{W}_r^{(l)}$, $\textbf{W}_i^{(l)}$, $\textbf{W}_j^{(l)},$ and  $\textbf{W}_k^{(l)} \in \mathbb{R}^{m\times n}$.
We now express the Hamilton product $\otimes$ between $\textbf{W}^{(l),Q}$ and $\boldsymbol{\mathsf{h}}_{\mathsf{u}}^{(l),Q}$ derived from Equation \ref{equa:halproduct_newform} as: 

\begin{equation}
\textbf{W}^{(l),Q}\otimes\boldsymbol{\mathsf{h}}_{\mathsf{u}}^{(l),Q} =  
\begin{bmatrix}
1\\
\boldsymbol{\mathsf{i}}\\
\boldsymbol{\mathsf{j}}\\
\boldsymbol{\mathsf{k}}\\
\end{bmatrix}^\top
\begin{bmatrix}
\textbf{W}_r^{(l)} & -\textbf{W}_i^{(l)} & -\textbf{W}_j^{(l)} & -\textbf{W}_k^{(l)}\\
\textbf{W}_i^{(l)} & \textbf{W}_r^{(l)} & -\textbf{W}_k^{(l)} & \textbf{W}_j^{(l)}\\
\textbf{W}_j^{(l)} & \textbf{W}_k^{(l)} & \textbf{W}_r^{(l)} & -\textbf{W}_i^{(l)}\\
\textbf{W}_k^{(l)} & -\textbf{W}_j^{(l)} & \textbf{W}_i^{(l)} & \textbf{W}_r^{(l)}
\end{bmatrix}
\begin{bmatrix}
\boldsymbol{\mathsf{h}}_{\mathsf{u},r}^{(l)}\\
\boldsymbol{\mathsf{h}}_{\mathsf{u},i}^{(l)}\\
\boldsymbol{\mathsf{h}}_{\mathsf{u},j}^{(l)}\\
\boldsymbol{\mathsf{h}}_{\mathsf{u},k}^{(l)}
\end{bmatrix}
\label{equa:quamul}
\end{equation}

Note that the quaternion components 
of $\textbf{W}^{(l),Q}$ 
are shared across the four quaternion components $\boldsymbol{\mathsf{h}}_{\mathsf{u},r}^{(l)}$, $\boldsymbol{\mathsf{h}}_{\mathsf{u},i}^{(l)}$, $\boldsymbol{\mathsf{h}}_{\mathsf{u},j}^{(l)}$, and  $\boldsymbol{\mathsf{h}}_{\mathsf{u},k}^{(l)}$.
Therefore, if we use any slight change in the input $\boldsymbol{\mathsf{h}}_{\mathsf{u}}^{(l),Q}$, we get an entirely different output, leading to a different performance.
This phenomenon is one of the crucial reasons why the Quaternion space provides highly expressive computations through the Hamilton product compared to the Euclidean and complex vector spaces \citep{parcollet2019survey}.
The phenomenon enforces the model to learn the potential relations within each hidden layer and between the different hidden layers, hence increasing the representation quality.
Furthermore, the four quaternion components $\textbf{W}_r^{(l)}$, $\textbf{W}_i^{(l)}$, $\textbf{W}_j^{(l)}$, and $\textbf{W}_k^{(l)}$ are shared when performing the Hamilton product; while in the Euclidean space, all the elements of the weight matrix are different parameter variables \citep{tay2019lightweight}. 
Thus, we can keep the same complexity and reduce the number of model parameters up to four times within the Quaternion space, similar to the parameter saving reported in \citep{parcollet2019quaternion,tay2019lightweight}.



\subsection{QGNN for node classification} 

Given a graph $\mathcal{G}$ where each node belongs to one of class labels, we are given the labels of a subset of $\mathcal{V}$. The node classification task is to predict the labels of remaining nodes.

We consider $\boldsymbol{\mathsf{h}}_\mathsf{v}^{(L),Q}$, which is the quaternion vector representation of node $\mathsf{v}$ at the {last} $L$-th layer.
To predict the label of node $\mathsf{v}$, we simply feed $\boldsymbol{\mathsf{h}}_\mathsf{v}^{(L),Q}$ to a prediction layer followed by a $\mathsf{softmax}$ layer as follows:
\begin{equation}
\boldsymbol{\mathsf{\hat{y}}}_{\mathsf{v}} = \mathsf{softmax}\left(\sum_{\mathsf{u} \in \mathcal{N}_\mathsf{v}\cup\left\{\mathsf{v}\right\}}a_{\mathsf{v},\mathsf{u}}\textbf{W}_1\textsc{Vec}\left(\boldsymbol{\mathsf{h}}_\mathsf{v}^{(L),Q}\right)\right) , \forall \mathsf{v} \in \mathcal{V}
\label{equa:nodecls}
\end{equation}
where \textsc{Vec}(.) denotes a concatenation of the four components of the quaternion vector. For example, 
\begin{equation}
\textsc{Vec}\left(\boldsymbol{\mathsf{h}}_\mathsf{v}^{(L),Q}\right) = \left[\boldsymbol{\mathsf{h}}_{\mathsf{v},r}^{(L)}\parallel\boldsymbol{\mathsf{h}}_{\mathsf{v},i}^{(L)}\parallel\boldsymbol{\mathsf{h}}_{\mathsf{v},j}^{(L)}\parallel\boldsymbol{\mathsf{h}}_{\mathsf{v},k}^{(L)}\right]
\label{equa:vectorization}
\end{equation}
We then learn the model parameters for the classification task by minimizing the cross-entropy loss function.

\subsection{QGNN for graph classification} 

Given a set of $M$ disjoint graphs $\left\{\mathcal{G}_m\right\}_{m=1}^M$ and their corresponding class labels $\left\{\mathsf{y}_m\right\}_{m=1}^M \subseteq \mathcal{Y}$, the graph classification task is to learn an embedding $\boldsymbol{\mathsf{e}}_{\mathcal{G}_m}$ for each entire graph $\mathcal{G}_m$ to predict its label $\mathsf{y}_m$.

Following \citep{xu2019powerful}, we obtain the quaternion embedding $\boldsymbol{\mathsf{e}}^Q_{\mathcal{G}}$ of the entire graph $\mathcal{G}$ as:
\begin{equation}
\boldsymbol{\mathsf{e}}^Q_{\mathcal{G}} = \sum_{\mathsf{v} \in \mathcal{V}}\boldsymbol{\mathsf{e}}^Q_\mathsf{v} = \sum_{\mathsf{v} \in \mathcal{V}}\left[\boldsymbol{\mathsf{h}}_\mathsf{v}^{(1),Q} \parallel \boldsymbol{\mathsf{h}}_\mathsf{v}^{(2),Q} \parallel ... \parallel \boldsymbol{\mathsf{h}}_\mathsf{v}^{(L),Q}\right] 
\end{equation}
To perform the task, we feed $\boldsymbol{\mathsf{e}}^Q_{\mathcal{G}}$ to a single fully-connected layer followed by a $\mathsf{softmax}$ layer as:
\begin{equation}
\boldsymbol{\mathsf{\hat{y}}}_{\mathcal{G}} = \mathsf{softmax}\left(\textbf{W}_2\textsc{Vec}\left(\boldsymbol{\mathsf{e}}^Q_{\mathcal{G}}\right) + \textbf{b}\right)
\label{equa:smgraphcls}
\end{equation}
where 
$\textsc{Vec}\left(\boldsymbol{\mathsf{e}}^Q_{\mathcal{G}}\right) = \left[\boldsymbol{\mathsf{e}}_{\mathcal{G},r} \parallel \boldsymbol{\mathsf{e}}_{\mathcal{G},i} \parallel \boldsymbol{\mathsf{e}}_{\mathcal{G},j} \parallel \boldsymbol{\mathsf{e}}_{\mathcal{G},k} \right]$.
We then also learn the model parameters by minimizing the cross-entropy loss function.

\subsection{QGNN for knowledge graph completion (SimQGNN)}

Knowledge graphs (KGs) can be viewed as directed multi-relational graphs to represent directional relationships between entities in the form of triples \textit{(head, relation, tail)} denoted as \textit{(h, r, t)}.
However, large KGs are still incomplete, i.e., missing a lot of valid triples \citep{West:2014}. 
To tackle this issue, research efforts have been made to predict whether a triple not in a knowledge graph is likely to be valid or not, which then helps to improve the graph completeness.
More specifically, many KG embedding models have been proposed to learn entity and relation embeddings and return a score for each triple \textit{(h, r, t)}, such that valid triples have higher scores than invalid ones \citep{bordes2011learning,NIPS2013_5071,NIPS2013_5028}. For example, the score of the valid triple (Melbourne, city\_Of, Australia) is higher than the score of the invalid one (Melbourne, city\_Of, Germany).

It is worth mentioning that several KG embedding approaches have been proposed to adapt GNNs for knowledge graph completion, e.g., R-GCN \citep{schlichtkrull2017modeling} and CompGCN \citep{vashishth2020compositionbased}. 
In general, these GNN-based models adopt an encoder-decoder architecture, wherein the encoder module aims to capture the relation-specific directions among entities, and then the decoder module adopts a score function to return the triple scores. Several examples of score functions are used in TransE \citep{NIPS2013_5071}, ComplEx \citep{Trouillon2016}, ConvE \citep{Dettmers2017}, and ConvKB \citep{Nguyen2018ConvKBfull}.

We note that the encoder modules used in the existing GNN-based KG embedding models focus on modifying the vanilla GNNs. For example, R-GCN modifies GCNs to associate entities with relation-specific transformation matrices, and CompGCN customizes GCNs to combine composition operations between entities and relations with relation-type specific weight matrices.
However, making these ``more complicated'' encoder modules is not completely effective at improving the performance of the existing GNN-based models.
This is probably because the directions between entities in the knowledge graph can be well-encoded by an effective score function.
In particular, the GNN-based models are outperformed by other conventional KG embedding models such as TuckER \citep{balazevic2019tucker} on benchmark datasets such as FB15k-237 \citep{toutanova-chen:2015:CVSC} for knowledge graph completion \citep{NIPS2013_5071}.

\begin{figure}[!ht]
	\centering
	\includegraphics[width=0.75\linewidth]{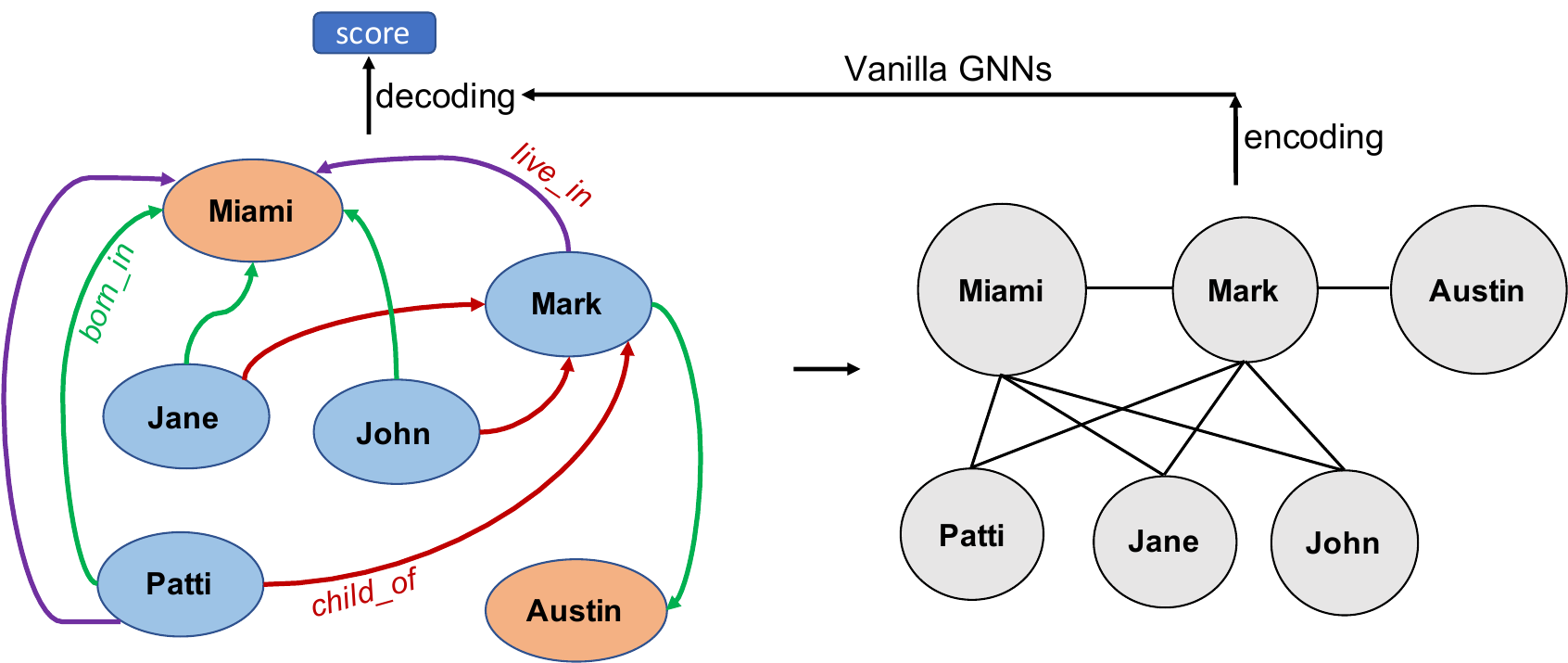}
    \caption{Illustration of adopting the vanilla GNNs on a single undirected graph $\mathcal{G}$ of entities built from a given knowledge graph $\mathit{G}$.}
    \label{fig:adoptingVanillaGNNs}
\end{figure}

To this end, we propose a new method of exploring the vanilla GNNs (such as GCN or our QGNN) directly on a single \textit{undirected} graph $\mathcal{G}$ of entities built from a given knowledge graph $\mathit{G}$, as illustrated in Figure \ref{fig:adoptingVanillaGNNs}, so that we construct a simple yet
effective adaptive encoder module.
As a result, we can easily adopt our QGNN on $\mathcal{G}$ as the encoder module as follows:
\begin{equation}
\boldsymbol{\mathsf{h}}_{\mathsf{e}}^{(l+1),Q} = \mathsf{g}\left(\sum_{\mathsf{e'} \in \mathcal{N}_\mathsf{e}\cup\left\{\mathsf{e}\right\}}a_{\mathsf{e},\mathsf{e'}}\textbf{W}^{(l),Q}\otimes\boldsymbol{\mathsf{h}}_{\mathsf{e'}}^{(l),Q}\right)
\end{equation}
where $\mathcal{N}_\mathsf{e}$ is the set of entity neighbors of entity node $\mathsf{e}$ in $\mathcal{G}$. 
We also adapt DistMult \citep{Yang2015} as a decoder module to compute the score of \textit{(h, r, t)} as:
\begin{equation}
f\left(h, r, t\right) = \left<\textsc{Vec}\left(\boldsymbol{\mathsf{h}}_h^{(L),Q}\right),\textsc{Vec}\left(\boldsymbol{v}_r^Q\right),\textsc{Vec}\left(\boldsymbol{\mathsf{h}}_t^{(L),Q}\right)\right> \nonumber
\end{equation}
where $\boldsymbol{\mathsf{h}}_h^{(L),Q}$ and $\boldsymbol{\mathsf{h}}_t^{(L),Q}$ are taken from the last QGNN layer; $\boldsymbol{\mathsf{h}}_h^{(0),Q}=\boldsymbol{v}_h^Q$ and $\boldsymbol{\mathsf{h}}_t^{(0),Q}=\boldsymbol{v}_t^Q$ are the quaternion embeddings of head entity $h$ and tail entity $t$, respectively;  $\boldsymbol{v}_r^Q$ is the quaternion embedding of relation $r$; and $\boldsymbol{v}_h^Q$, $\boldsymbol{v}_r^Q$, and $\boldsymbol{v}_t^Q$ are learned during training. \textsc{Vec}(.) denotes a concatenation of the four components of the quaternion vector as mentioned in Equation \ref{equa:vectorization}; and $\left<\right>$ denotes a multiple-linear dot product. 

Similarity, we can also adopt GCN as the encoder module.
We name our QGNN-based KG embedding model as \textbf{SimQGNN} and its variant using GCN as \textbf{SimRGCN}.

We then employ the Adam optimizer \citep{kingma2014adam} to train our proposed model by minimizing the binary cross-entropy loss function \citep{Dettmers2017}  as:
\begin{eqnarray}
\mathcal{L} = -\sum_{\substack{(h, r, t) \in \{\mathit{G} \cup \mathit{G}'\}}}\left(\textsc{l}_{(h, r, t)}\log\left(p_{(h, r, t)}\right)\right. + 
\left.\left(1 - \textsc{l}_{(h, r, t)}\right)\log\left(1 - p_{(h, r, t)}\right)\right)
\label{equal:losscrossentropy}
\end{eqnarray}
\begin{equation*}
\text{in which, } \textsc{l}_{(h, r, t)} = \left\{ 
  \begin{array}{l}
1 \ \ \ \text{for } (h, r, t)\in\mathit{G}\\
0 \ \ \ \text{for } (h, r, t)\in\mathit{G}'
  \end{array} \right.
\end{equation*}
where $p_{(h, r, t)} = \mathsf{sigmoid}\left(f(h, r, t)\right)$. $\mathit{G}$ and $\mathit{G}'$ are collections of valid and invalid triples, respectively.

\section{Experimental results}

\subsection{QGNN for graph classification}

\paragraph{Datasets} 
We evaluate our QGNN on well-known datasets consisting of three social network datasets (i.e., COLLAB, IMDB-B and IMDB-M) and four bioinformatics datasets (i.e., DD, MUTAG, PROTEINS, and PTC) \citep{yanardag2015deep}. 
The social network datasets do not have available node features; thus, we follow \citep{niepert2016learning,zhang2018end,xu2019powerful} to use node degrees as features on these datasets.

\paragraph{Evaluation protocol}
We follow \citep{xu2019powerful,maron2019provably,seo2019discriminative,Chen2019ArePG} to use the {same} data splits and the {same} 10-fold cross-validation scheme to calculate the classification performance for a fair comparison.
We compare our QGNN with up-to-date strong baselines and report the baseline results reported either in the original papers or in \citep{verma2018graph,xinyi2019capsule,Chen2019ArePG,seo2019discriminative,xu2019powerful}.

\paragraph{Training protocol}
We usually evaluate the proposed models on benchmark datasets where the Euclidean feature vectors $\boldsymbol{\mathsf{h}}_\mathsf{v}$ are typically given and pre-fixed. 
Hence we set the same $\boldsymbol{\mathsf{h}}_\mathsf{v}$ to the four components of  $\boldsymbol{\mathsf{h}}_\mathsf{v}^{(0),Q}$ as:
$\boldsymbol{\mathsf{h}}_{\mathsf{v},r}^{(0)} = \boldsymbol{\mathsf{h}}_{\mathsf{v},i}^{(0)} = \boldsymbol{\mathsf{h}}_{\mathsf{v},j}^{(0)} = \boldsymbol{\mathsf{h}}_{\mathsf{v},k}^{(0)} = \boldsymbol{\mathsf{h}}_\mathsf{v}$
This is similar to the way applying the Quaternion space on grayscale images \citep{zhu2018quaternion}.
We vary the number of hidden layers in \{1, 2, 3, 4, 5\}, and the hidden size (i.e., the number of quaternions in the hidden layers) in \{8, 16, 32, 64, 128\}.
We set the batch size to 4 and use the Adam optimizer \citep{kingma2014adam} with the initial learning rate in $ \left\{5e^{-5}, 1e^{-4}, 5e^{-4}, 1e^{-3}\right\}$.
We run up to 100 epochs to evaluate our trained model.

\paragraph{Experimental results}

\begin{table*}[!ht]
\caption{Graph classification accuracies (\%). The best scores are in \textbf{bold}.
} 
\centering
\resizebox{15.5cm}{!}{
\def\arraystretch{1.2}
\begin{tabular}{lccccccc}
\hline
{\bf Model} &  \textbf{COLLAB} &  \textbf{IMDB-B} &  \textbf{IMDB-M} &  \textbf{DD} & \textbf{PROTEINS} & \textbf{MUTAG} & \textbf{PTC}\\
\hline
PSCN \citeyearpar{niepert2016learning} & 72.60 $\pm$ 2.15 & 71.00 $\pm$ 2.29 & 45.23 $\pm$ 2.84 & 77.12 $\pm$ 2.41 & 75.89 $\pm$ 2.76 & \textbf{92.63 $\pm$ 4.21} & 62.29 $\pm$ 5.68\\
GCN \citeyearpar{kipf2017semi} & 79.00 $\pm$ 1.80 & 74.00 $\pm$ 3.40 & 51.90 $\pm$ 3.80 & -- & 76.00 $\pm$ 3.20 & 85.60 $\pm$ 5.80 & 64.20 $\pm$ 4.30\\
GraphSAGE \citeyearpar{hamilton2017inductive} & 79.70 $\pm$ 1.70 & 72.40 $\pm$ 3.60 & 49.90 $\pm$ 5.00 & 65.80 $\pm$ 4.90 & 65.90 $\pm$ 2.70 & 79.80 $\pm$ 13.9 & -- \\
GAT \citeyearpar{velickovic2018graph} & 75.80 $\pm$ 1.60 & 70.50 $\pm$ 2.30 & 47.80 $\pm$ 3.10 & -- & 74.70 $\pm$ 2.20 & 89.40 $\pm$ 6.10 & 66.70 $\pm$ 5.10 \\
GCAPS \citeyearpar{verma2018graph} & 77.71 $\pm$ 2.51 & 71.69 $\pm$ 3.40 & 48.50 $\pm$ 4.10 & 77.62 $\pm$ 4.99 & 76.40 $\pm$ 4.17 & -- & 66.01 $\pm$ 5.91\\
DGCNN \citeyearpar{zhang2018end} & 73.76 $\pm$ 0.49 & 70.03 $\pm$ 0.86 & 47.83 $\pm$ 0.85 & 79.37 $\pm$ 0.94 & 75.54 $\pm$ 0.94 & 85.83 $\pm$ 1.66 & 58.59 $\pm$ 2.47\\
CapsGNN \citeyearpar{xinyi2019capsule} & 79.62 $\pm$ 0.91 & 73.10 $\pm$ 4.83 & 50.27 $\pm$ 2.65 & 75.38 $\pm$ 4.17 & 76.28 $\pm$ 3.63 & 86.67 $\pm$ 6.88 & -- \\
GIN-0 \citeyearpar{xu2019powerful} & 80.20 $\pm$ 1.90  & 75.10 $\pm$ 5.10 & 52.30 $\pm$ 2.80 & -- & 76.20 $\pm$ 2.80 & {89.40 $\pm$ 5.60} & 64.60 $\pm$ 7.00\\
IEGN \citeyearpar{maron2019invariant} & 77.92 $\pm$ 1.70 & 71.27 $\pm$ 4.50 & 48.55 $\pm$ 3.90 & -- & 75.19 $\pm$ 4.30 & 84.61 $\pm$ 10.0 & 59.47 $\pm$ 7.30\\
PPGN \citeyearpar{maron2019provably} & 81.38 $\pm$ 1.42 & 73.00 $\pm$ 5.77 & 50.46 $\pm$ 3.59 & -- & 77.20 $\pm$ 4.73 & 90.55 $\pm$ 8.70 &  66.17 $\pm$ 6.54\\
GFN \citeyearpar{Chen2019ArePG} & \textbf{81.50 $\pm$ 2.42} & 73.00 $\pm$ 4.35 & 51.80 $\pm$ 5.16 & 78.78 $\pm$ 3.49 & 76.46 $\pm$ 4.06 & {90.84 $\pm$ 7.22} & --\\
\hline
\textbf{QGNN} & 81.36 $\pm$ 1.31 & \textbf{77.56 $\pm$ 2.45} & \textbf{53.78 $\pm$ 3.83} & \textbf{79.92 $\pm$ 3.54} & \textbf{78.47 $\pm$ 3.30} & {92.59 $\pm$ 3.59} & \textbf{69.92 $\pm$ 2.59}\\
\hline
\end{tabular}
}
\label{tab:expresults_sup}
\end{table*}

Table \ref{tab:expresults_sup} presents the accuracy results of our QGNN and other up-to-date baselines.
In general, our QGNN produces state-of-the-art accuracies on most datasets; hence this demonstrates a notable impact of our model. 
Especially, QGNN outperforms the existing baseline models and produces state-of-the-art accuracies 
on IMDB-B, IMDB-M, DD, PROTEINS, and PTC, respectively.
QGNN also obtains competitive accuracies 
on COLLAB and MUTAG, respectively. 
Furthermore, compared to GCN, the obtained results demonstrate the effectiveness of QGNN to generalize GCN within the Quaternion space for the graph classification task. 



\subsection{QGNN for node classification}
\label{sec:nodeclsprotocol}

\paragraph{Datasets} We use three well-known benchmark datasets consisting of \textsc{Cora, Citeseer} \citep{sen2008collective} and \textsc{Pubmed} \citep{namata:mlg12} that are citation networks.


\paragraph{Evaluation protocol} 
As mentioned in \citep{fey2019fast,Pei2020GeomGCN}, the experimental setup used in \citep{kipf2017semi,velickovic2018graph} is not fair to show the effectiveness of existing GNN models when only using one fixed data split of training, validation, and test sets from \citep{Yang:2016planetoid}.
Therefore, for a fair comparison, we use the {same} 10 random data splits used in \citep{Pei2020GeomGCN}, where each data split consists of 60\%, 20\%, 20\% numbers of nodes, equally distributed for each node class, for training, validation, and testing, respectively.
We also follow \citep{Pei2020GeomGCN} to report the mean classification accuracy on the test sets over the 10 data splits.

\paragraph{Training protocol}

We also we set the same $\boldsymbol{\mathsf{h}}_\mathsf{v}$ to the four components of  $\boldsymbol{\mathsf{h}}_\mathsf{v}^{(0),Q}$.
The architecture used in \citep{Pei2020GeomGCN} is a 2-layer GCN, wherein the hidden sizes are 16 for \textsc{Cora} and \textsc{Citeseer}, and 64 for \textsc{Pubmed}. 
Hence, we construct one QGNN layer followed by a prediction layer and then a $\mathsf{softmax}$ layer (referring to our QGNN for node classification as mentioned in Equation \ref{equa:nodecls}). 
We use the corresponding hidden sizes of 4 for \textsc{Cora} and \textsc{Citeseer}, and 16 on \textsc{Pubmed}.
We also set the same Adam initial learning rate to 0.05, and the same number of epochs to 100 for both \textsc{Cora} and \textsc{Citeseer}; while they are 0.1 and 200 respectively for \textsc{Pubmed}.
We also provide the accuracy results of Hyperbolic Graph Convolutional Neural Networks (HGCN) \citep{chami2019hyperbolic} following these evaluation and training protocols. 

\paragraph{Experimental results}

\begin{table}[!ht]
\centering
\caption{Node classification accuracies (\%).}
\def\arraystretch{1.1}
\setlength{\tabcolsep}{0.5em}
\begin{tabular}{lcccc}
\hline
{\bf Dataset} & \textbf{GAT} & \textbf{GCN} & \textbf{HGCN} & \textbf{QGNN}\\
\hline
\textsc{Cora} & 86.37 & 85.77 & 86.09 & \textbf{87.48} \\
\textsc{Citeseer} & 74.32 & 73.68 & 74.84 & \textbf{76.03}\\
\textsc{Pubmed} & 87.62 & \textbf{88.13}  & 87.13 & 87.65\\
\hline 
\end{tabular}
\label{tab:nodeclsresults}
\end{table}

Table \ref{tab:nodeclsresults} presents the node classification accuracies, where the results of GCN and GAT are also taken from \citep{Pei2020GeomGCN}.
Our QGNN achieves the accuracies of 87.48\%, 76.03\%, and 87.65\% on \textsc{Cora}, \textsc{Citeseer}, and \textsc{Pubmed} respectively.
In particular, QGNN outperforms GCN, GAT and HGCN on \textsc{Cora} and \textsc{Citeseer}, and produces competitive results on \textsc{Pubmed}, e.g., QGNN obtains absolute gains of 1.39\%, 1.19\%, and 0.52\% over HGCN on \textsc{Cora}, \textsc{Citeseer}, and \textsc{Pubmed}, respectively.



\subsection{QGNN for knowledge graph completion (SimQGNN)}
\label{subsec:kglp}

We evaluate our proposed method for the knowledge graph completion task, i.e., link prediction \citep{NIPS2013_5071}, which aims to predict a missing entity given a relation with another entity, e.g., inferring a head entity $h$ given $(?, r, t)$ or inferring a tail entity $t$ given $(h, r, ?)$. 
The results are computed by ranking the scores returned by the score function $f$ on triples in the test set.

\paragraph{Datasets}
\citet{safavi2020codex} point out issues with existing KG completion datasets and thus present three new and more appropriately difficult benchmark datasets CoDEx-S, CoDEx-M, and CoDEx-L.
These three open-domain CoDEx datasets are derived from Wikidata and Wikipedia to cover more diverse and interpretable content and make a more challenging prediction task. Therefore, we use these new datasets in our experiments.


\paragraph{Evaluation protocol}
Following \citet{NIPS2013_5071}, for each valid test triple $(h, r, t)$, we replace either $h$ or $t$ by each of all other entities to create a set of corrupted triples.
We also use the ``{Filtered}'' setting protocol \citep{NIPS2013_5071}.
We rank the valid test triple and corrupted triples in descending order of their scores to calculate mean reciprocal rank (MRR) and Hits@$10$.
The final scores on the test set are reported for the model which obtains the highest Hits@10 on the validation set.

\paragraph{Training protocol}

We set the batch size to 1024 and employ the Adam optimizer \citep{kingma2014adam} to train our model up to 4,000 epochs on CoDEx-S and CoDEx-M, and 2,000 epochs on CoDEx-L.
We set the same dimension value for both the quaternion embedding size and the hidden size, wherein we vary the dimension value in $\{32, 64, 128, 256\}$.
We also vary the number of hidden layers in $\{1, 2, 3\}$ and the Adam initial learning rate in $\left\{1e^{-4}, 5e^{-4}, 1e^{-3}, 5e^{-3}, 1e^{-2}\right\}$.
We use grid search to select the best model checkpoints, wherein we compute the Hits@10 scores after each training epoch on the CoDEx-S and CoDEx-M validation sets and after each 5 training epochs on the CoDEx-L validation set.

For other baseline models including DistMult \citep{Yang2015}, 
SACN \citep{shang2019end}, R-GCN \citep{schlichtkrull2017modeling}, and CompGCN \citep{vashishth2020compositionbased}, we apply the same evaluation protocol. The training protocol is the same w.r.t. the optimizer, the hidden layers, the initial learning rate values, and the number of training epochs. 

\paragraph{Experimental results}

\begin{table}[!ht]
\caption{Experimental results on the {test} sets. Hits@10 (H@10) is reported in \%. 
The best scores are in {bold}, while the second best scores are in {underline}.
$*$ denotes the results taken from \citep{safavi2020codex}, while ${\clubsuit}$ denotes our own results for other baseline models. 
We get an out-of-memory for SACN on the large dataset CoDEx-L.
}
\centering
\setlength{\tabcolsep}{0.4em}
\def\arraystretch{1.1}
\begin{tabular}{l|cc|cc|cc}
\hline
\multirow{2}{*}{\bf Method} & \multicolumn{2}{c|}{\bf CoDEx-S} & \multicolumn{2}{c|}{\bf CoDEx-M} & \multicolumn{2}{c}{\bf CoDEx-L} \\
\cline{2-7} 
& MRR   & H@10  & MRR   & H@10  & MRR   & H@10 \\
\hline
TransE$^*$  & 0.354 & 63.4 & 0.303 & 45.4 & 0.187 & 31.7 \\
ComplEx$^*$ & \bf 0.465 & \underline{64.6} & \bf 0.337 & \underline{47.6} & 0.294 & 40.0 \\
ConvE$^*$   & \underline{0.444} & 63.5 & 0.318 & 46.4 & 0.303 & 42.0 \\
TuckER$^*$  & \underline{0.444} & 63.8 & \underline{0.328} & 45.8 & \underline{0.309} & {43.0} \\
\hline
DistMult$^{\clubsuit}$& 0.435 & {64.6} & 0.320 & 46.8 & 0.307 & 42.7 \\
\hdashline
R-GCN$^{\clubsuit}$   & 0.275 & 53.3 & 0.124 & 24.1 & 0.073 & 14.2 \\
SACN$^{\clubsuit}$    & 0.374 & 59.4 & 0.294 & 44.3 & -- & --  \\
CompGCN$^{\clubsuit}$ & 0.395 & 62.1 & 0.312 & 45.7 & 0.304 & 42.8 \\
\hline
\textbf{SimRGCN} & 0.427 & \underline{64.7} & 0.322 & 47.5 & 0.307 & \underline{43.2}\\
\hdashline
\textbf{SimQGNN}  & 0.435 & \textbf{65.2} & 0.323 & \textbf{47.7} & \textbf{0.310} & \textbf{43.7} \\
\hline
\end{tabular}
\label{tab:resultsCoDEx}
\end{table}

We report the  results on the CoDEx datasets for our proposed SimQGNN and SimRGCN, and other baseline models in Table \ref{tab:resultsCoDEx}.
SimQGNN performs better than the more complicated models R-GCN, SACN, and CompGCN, e.g., SimQGNN obtains absolute Hits@10 improvements of 3.1\%, 2.0\%, and 0.9\% over CompGCN on CoDEx-S, CoDEx-M, and CoDEx-L, respectively.
In general, our SimQGNN obtains the highest Hits@10 scores and achieves competitive MRR scores on CoDEx-S and CoDEx-M. Furthermore, SimQGNN produces the highest MRR and Hits@10 scores on the large and more challenging dataset CoDEx-L.



It is worth mentioning that our SimRGCN outperforms R-GCN, SACN, and CompGCN; hence this clearly shows the effectiveness of our proposed method of exploring the vanilla GNNs directly on the single undirected graph $\mathcal{G}$ of entities built from the given knowledge graph $\mathit{G}$.
Besides, it is noted that, compared to SimQGNN, the results degrade for SimRGCN, implying the advantage of our QGNN over GCN.

\section{Conclusion}
\label{sec:conclusion}

We propose Quaternion Graph Neural Networks (QGNN) to learn graph representations within the Quaternion space. 
Our proposed QGNN obtains state-of-the-art accuracies on well-known benchmark datasets for node classification and graph classification.
Furthermore, regarding knowledge graphs, our QGNN-based embedding model, named SimQGNN, outperforms the existing models and produces state-of-the-art results on three benchmark datasets CoDEx for knowledge graph completion.\footnote{
We introduce Gated Quaternion Graph Neural Networks (GQGNN) to learn graph representations in Appendix \ref{app:appendixgqnn}. Our GQGNN-based text classification model, called TextQGNN, performs better than the existing models on well-known benchmark datasets for inductive text classification.
}

\bibliography{references_kg,references}

\appendix

\section{Gated Quaternion Graph Neural Networks (GQGNN)}
\label{app:appendixgqnn}

We propose Gated Quaternion Graph Neural Networks (GQGNN), which can be seen as a generalization of Gated GNNs \citep{li2015gated} within the Quaternion space as follows:
\begin{eqnarray}
\textbf{a}_{\mathsf{v}}^{(l+1),Q} &=& \sum_{\mathsf{u} \in \mathcal{N}_\mathsf{v}}a_{\mathsf{v},\mathsf{u}}\boldsymbol{\mathsf{h}}_{\mathsf{u}}^{(l),Q} \nonumber\\
\textbf{z}_{\mathsf{v}}^{(l+1),Q} &=& \sigma\left(\textbf{W}^{z,Q}\otimes\textbf{a}_{\mathsf{v}}^{(l+1),Q} + \textbf{U}^{z,Q}\otimes\boldsymbol{\mathsf{h}}_{\mathsf{v}}^{(l),Q}\right) \nonumber \\
\textbf{r}_{\mathsf{v}}^{(l+1),Q} &=& \sigma\left(\textbf{W}^{r,Q}\otimes\textbf{a}_{\mathsf{v}}^{(l+1),Q} + \textbf{U}^{r,Q}\otimes\boldsymbol{\mathsf{h}}_{\mathsf{v}}^{(l),Q}\right) \nonumber \\
\widetilde{\boldsymbol{\mathsf{h}}_{\mathsf{v}}^{(l+1),Q}} &=& \mathsf{g}\left(\textbf{W}^{o,Q}\otimes\textbf{a}_{\mathsf{v}}^{(l+1),Q} + \textbf{U}^{o,Q}\otimes\left(\textbf{r}_{\mathsf{v}}^{(l+1),Q}\odot\boldsymbol{\mathsf{h}}_{\mathsf{v}}^{(l),Q}\right)\right) \nonumber \\
\boldsymbol{\mathsf{h}}_{\mathsf{v}}^{(l+1),Q} &=& \left(1 - \textbf{z}_{\mathsf{v}}^{(l+1),Q}\right)\odot\boldsymbol{\mathsf{h}}_{\mathsf{v}}^{(l),Q} + \textbf{z}_{\mathsf{v}}^{(l+1),Q}\odot\widetilde{\boldsymbol{\mathsf{h}}_{\mathsf{v}}^{(l+1),Q}}
\end{eqnarray}
where $\textbf{z}$ and $\textbf{r}$ are the update and reset gates; $\sigma$ is the sigmoid function; $\odot$ is element-wise multiplication; and $\mathsf{g}$ is a nonlinear activation function (such as $\mathsf{ReLU}$).

\subsection{GQGNN for inductive text classification (TextQGNN)}
We follow \citep{zhang2020every} to build a graph $\mathcal{G}$ for each textual document by representing unique words as nodes and co-occurrences between words as edges. 
We then employ our proposed GQGNN to update node representations.
To obtain the graph embedding $\boldsymbol{\mathsf{e}}^Q_{\mathcal{G}}$, we follow \citep{anonymous2021regvd} to define a readout function as:
\begin{eqnarray}
\boldsymbol{\mathsf{e}}^Q_\mathsf{v} &=& \sigma\left(\textbf{w}^\mathsf{T}\textsc{Vec}\left(\boldsymbol{\mathsf{h}}_\mathsf{v}^{(L),Q}\right) + \mathsf{b}\right)\odot\mathsf{g}\left(\textbf{W}^Q\otimes\boldsymbol{\mathsf{h}}_{\mathsf{v}}^{(L),Q}\right) \\
\boldsymbol{\mathsf{e}}^Q_{\mathcal{G}} &=& \sum_{\mathsf{v} \in \mathcal{V}}\boldsymbol{\mathsf{e}}^Q_\mathsf{v}\odot\mathsf{max\_pooling}\left\{\boldsymbol{\mathsf{e}}^Q_\mathsf{v}\right\}_{\mathsf{v} \in \mathcal{V}}
\end{eqnarray}
where $\sigma\left(\textbf{w}^\mathsf{T}\textsc{Vec}\left(\boldsymbol{\mathsf{h}}_\mathsf{v}^{(L),Q}\right) + \mathsf{b}\right)$ acts as soft attention mechanisms over nodes.
Finally, we also follow Equation \ref{equa:smgraphcls} to perform the task. We name our GQGNN-based text classification model as \textbf{TextQGNN}.

\subsection{Experimental setup and results}
\label{sec:textclsprotocol}

\paragraph{Datasets} 
We follow \citep{yao2019graph,zhang2020every} to use four benchmarks -- R8, R52, \textsc{Ohsumed}, and MR.

\paragraph{Training protocol} 
We also follow \citep{zhang2020every} to use random vectors or pre-trained Glove \citep{pennington2014glove} with the dimension size of 300 to initialize the Euclidean feature vectors. 
Then, for each Euclidean feature vector, we simply split it into four parts to create four corresponding components of a Quaternion feature vector with the dimension size of 75.
We follow \citep{yao2019graph,zhang2020every} to construct 2-layer steps. 
We set the quaternion hidden size to 96 and vary the learning rate in $ \left\{1e^{-4}, 5e^{-4}, 1e^{-3}, 5e^{-3}\right\}$. 
We use the Adam optimizer to train the model up to 150 epochs to evaluate our trained model.

\paragraph{Evaluation protocol} For each dataset, we report the mean accuracy and standard deviation over 10 times, wherein for each time, we randomly sample 10\% text from the training set to construct the validation set for hyper-parameter turning.

\begin{table}[!ht]
\centering
\caption{Text classification accuracies (\%) on the test sets. Some baseline results are taken from \citep{yao2019graph}.
We report the new results of TextING for the model which obtains the highest accuracy on the validation set for a fair comparison.}
\def\arraystretch{1.2}
\resizebox{11cm}{!}{
\setlength{\tabcolsep}{0.85em}
\begin{tabular}{lcccc}
\hline 
{\bf Dataset} &  \textbf{MR} & \textbf{R8} & \bf R52 & \bf \textsc{Ohsumed} \\
\hline
Bi-LSTM & 77.68 $\pm$ 0.86 & 96.31 $\pm$ 0.33 & 90.54 $\pm$ 0.91 & 49.27 $\pm$ 1.07 \\
fastText & 75.14 $\pm$ 0.20 & 96.13 $\pm$ 0.21 & 92.81 $\pm$ 0.09 & 57.70 $\pm$ 0.49 \\
TextGCN & 76.74 $\pm$ 0.20 & 97.07 $\pm$ 0.10 & 93.56 $\pm$ 0.18 & 68.36 $\pm$ 0.56 \\
TextING & 78.86 $\pm$ 0.26 & 96.90 $\pm$ 0.23 & 93.34 $\pm$ 0.24 & 69.72 $\pm$ 0.30 \\ 
\hline
\textbf{TextQGNN} & 78.93 $\pm$ 0.29 & 97.02 $\pm$ 0.28 & 94.45 $\pm$ 0.35 & 69.93 $\pm$ 0.31\\
\hline
\end{tabular}
}
\label{tab:textclsresults}
\end{table}

\paragraph{Experimental results}
Table \ref{tab:textclsresults} presents the text classification accuracies of our TextQGNN and the baselines. In general, our TextQGNN works better than the baselines on the benchmark datasets.

\end{document}